\documentclass[lettersize,journal]{IEEEtran}
\usepackage{amsmath,amsfonts}
\usepackage{algorithmic}
\usepackage{algorithm}
\usepackage{array}
\usepackage[caption=false,font=normalsize,labelfont=sf,textfont=sf]{subfig}
\usepackage{textcomp}
\usepackage{stfloats}
\usepackage{url}
\usepackage{verbatim}
\usepackage{graphicx}
\usepackage{cite}
\usepackage[T1]{fontenc}
\usepackage{bbding}
\usepackage{multirow}
\usepackage{booktabs}
\begin{document}

\title{Joint CNN and Transformer Network via weakly supervised Learning for efficient crowd counting}

\author{
	Fusen Wang,
	Kai Liu,
	Fei Long,
	Nong Sang,~\IEEEmembership{Member,~IEEE,}
	Xiaofeng Xia,
	Jun Sang
	\thanks{Fusen Wang and Kai Liu and Fei Long and Xiaofeng Xia and Jun Sang are with the School of Big Data \& Software Engineering, Chongqing University, Chongqing 401331, China (201924131014@cqu.edu.cn; kailiu@cqu.edu.cn; 201924132039@cqu.edu.cn; nsang@hust.edu.cn; xiaxiaofeng@cqu.edu.cn; jsang@cqu.edu.cn). Jun Sang is the corresponding author.}
	\thanks{Nong Sang is with the School of Artificial Intelligence and Automation, Huazhong University of Science and Technology, Wuhan 430074, China.}
}



\maketitle

\begin{abstract}
Currently, for crowd counting, the fully supervised methods via density map estimation are the mainstream research directions.
However, such methods need location-level annotation of persons in an image, which is time-consuming and laborious.
Therefore, the weakly supervised method just relying upon the count-level annotation is urgently needed.
Since CNN is not suitable for modeling the global context and the interactions between image patches, crowd counting with weakly supervised learning via CNN generally can not show good performance.
The weakly supervised model via Transformer was sequentially proposed to model the global context and learn contrast features.
However, the transformer directly partitions the crowd images into a series of tokens, which may not be a good choice due to each pedestrian being an independent individual, and the parameter number of the network is very large.
Hence, we propose a Joint CNN and Transformer Network (JCTNet) via weakly supervised learning for crowd counting in this paper.
JCTNet consists of three parts: CNN feature extraction module (CFM), Transformer feature extraction module (TFM), and counting regression module (CRM).
In particular, the CFM extracts crowd semantic information features, then sends their patch partitions to TRM for modeling global context, and CRM is used to predict the number of people.
Extensive experiments and visualizations demonstrate that JCTNet can effectively focus on the crowd regions and obtain superior weakly supervised counting performance on five mainstream datasets. 
The number of parameters of the model can be reduced by about 67\%$\sim$73\% compared with the pure Transformer works.
We also tried to explain the phenomenon that a model constrained only by count-level annotations can still focus on the crowd regions. 
We believe our work can promote further research in this field.
\end{abstract}

\begin{IEEEkeywords}
Crowd counting, CNN, Transformer, Weakly supervised learning, Count-level annotations.
\end{IEEEkeywords}

\section{Introduction}
Crowd Counting based on deep learning is a hot topic due to its significant role in many applications, such as crowd tracking, public transportation, transfer counting, etc. The purpose of crowd counting is to obtain the counting of each image.
The current mainstream works \cite{MCNN,CrowdNet,DADNet,CAN,SFANet,schroder2018optical,SFCN} usually adopt the learning method of location-level supervision and have achieved significant precision promotion.
However, these works require location-level annotation of each person, which has some weaknesses: (1) Collecting the location-level person annotations can become expensive and laborious in a large crowd scenario, such as, in NWPU-Crowd \cite{NWPU}, the image of the largest scene contains 20,033 person head annotations. 
(2) Due to the gap between the training and inferencing phases in crowd counting, location-level annotations are only used for the former and redundant for the latter \cite{Transcrowd}.
Therefore, weakly supervised learning with counting-level annotations for crowd counting is essential.
In addition, MATT \cite{MATT} pointed out that the total count of objects usually can be economically obtained in many realistic scenes.

\begin{figure}[!t]
	\centering
	\includegraphics[width=2.5in]{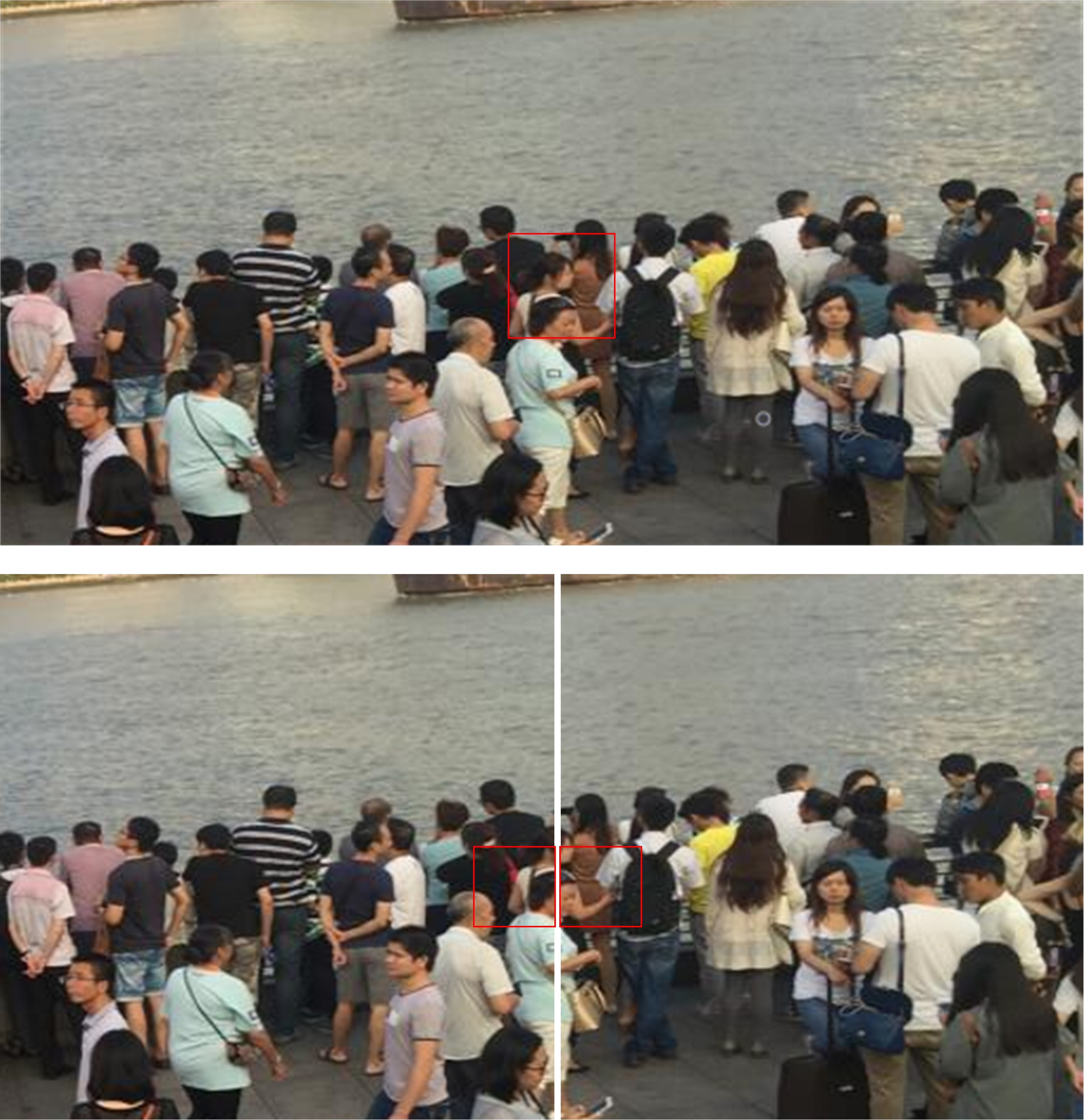}
	\caption{Patch Partition in the input image. In the middle, several person heads are divided into two parts, which is unreasonable and may lead to overestimation or underestimation.}
	\label{fig_1}
\end{figure}

In recent works, the counting performance of the weakly supervised model via CNN \cite{MATT,HACNN} is relatively poor due to convolution kernels' limited receptive field and failure in modeling the global context information, which plays a vital role in crowd counting \cite{CAN,BCCT}.
Besides, CNN can not establish the interactions between image patches, so it is incapable of learning the contrast features between crowd and background.
Compared with the CNN, the Transformer can adequately learn the contrast features between different image patches due to the self-attention mechanism in image classification, segmentation \cite{VIT,SwinTransformer,Twins}, etc.
This features can be significantly used to distinguish crowds from background areas.
Hence, some researchers developed transformer architecture to effectively capture crowd global context information and model long-range dependency \cite{Transcrowd,CCTrans,BCCT}, which all achieved superior counting performance.
However, it is unreasonable for the Transformer to directly divide the input images into a series of patches as tokens.
As shown in Fig. 1, the input image is divided into two patches, and several person heads are split into two parts, which may cause overestimation or underestimation due to each pedestrian being an independent individual.
Besides, the pure transformer model needs more parameters to achieve superior performance due to the lack of CNN's inductive bias ability \cite{VIT,SwinTransformer,Twins}.
To sum up, CNN and Transformer may be complementary.

\begin{figure*}[!t]
	\centering
	\includegraphics[width=5.0in]{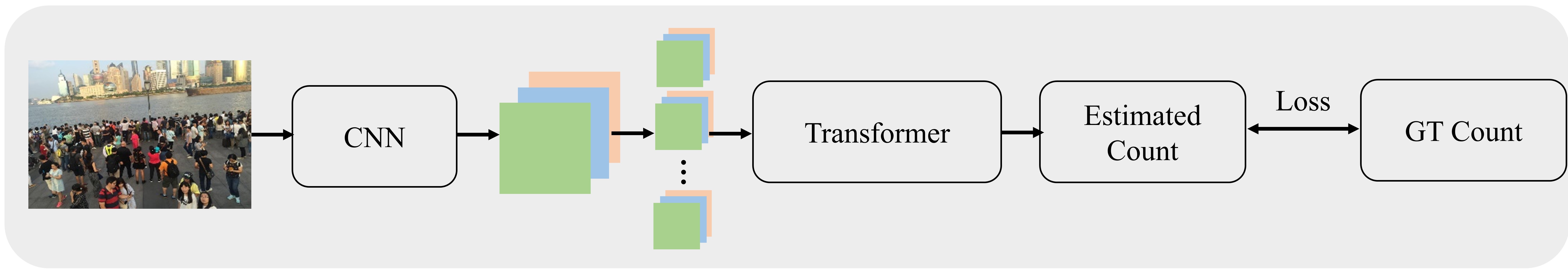}
	\caption{The overview of the proposed Weakly supervised counting via Joint CNN and Transformer, namely JCTNet. The CNN is used to extract crowd semantic information and the Transformer is utilized to model global context and learn contrast features for improving the final counting performance.}
	\label{fig_2}
\end{figure*}

In this paper, we propose a Joint CNN and Transformer Network (JCTNet) for weakly supervised counting, as shown in Fig. 2, which effectively alleviates the above issues and achieves better counting performance with fewer parameters.
A detailed pipeline diagram is shown in Fig. 3. 
JCTNet consists of three modules: CNN feature extraction module (CFM), Transformer feature extraction module (TFM), and counting regression module (CRM).
Specifically, CFM utilizes VGG16-BN \cite{VGG} first ten layers to extract crowd semantic features, which can relieve the issue of dividing the input images.
Similar to \cite{SwinTransformer}, TFM includes four modified Swin Transformer blocks (MSTB) for local attention and cross-window interaction.
The semantic features from CFM are transmitted into TFM to capture global context and model long-range dependency.
Finally, the token features from TFM are reshaped into three dimensions and utilized to estimate the number of people by CRM.

To summarize, the main contributions of our work are outlined as follows:
\begin{itemize}
	\item[(1)] We propose a Joint CNN and Transformer Network (JCTNet) with complementary learning for weakly supervised crowd counting, which only relies on count-level annotations to reduce annotation costs. In addition, this joint learning can make up for the deficiencies between CNN and Transformer for further improving the counting performance.
	\item[(2)] On five mainstream datasets (ShanghaiTech Part A/B, UCF-CC-50, UCF-QNRF, NWPU-Crowd), extensive experiments demonstrate that JCTNet obtains superior counting performance.
	\item[(3)] The proposed JCTNet has fewer parameters and achieves lower counting errors compared with the pure transformer works. As shown in Tabel 1, JCTNet obtains lower counting errors (MAE of 62.8) on the ShanghaiTech Part A dataset \cite{MCNN} and only employs 33\% and 27\% of the parameters of TransCrowd \cite{Transcrowd} and CCTrans \cite{CCTrans}, respectively. In addition, we analyze why this model constrained only by count-level annotations can focus on crowd regions in Section 5 and hope this work could encourage further research in the weakly supervised crowd counting.
\end{itemize}

\begin{table*}[ht]
	\centering
	\caption{Counting errors of several weakly-supervised crowd counting methods on ShanghaiTech Part A.}
	\label{Table 1}
	\begin{tabular}{|c|c|c|c|c|c|}
		\hline
	    Method   & \multicolumn{1}{c|}{Publication} & \multicolumn{1}{c|}{Model}   & MAE  & MSE   & \multicolumn{1}{c|}{Parameters(M)} \\ \hline
		Yang et al. \cite{yang2020weakly}*   & ECCV20  & CNN  & 104.6  & 145.2	& ---	  \\
		MATT \cite{MATT}*   & PR21 & CNN & 80.1  & 129.4  & ---	 \\
		TransCrowd \cite{Transcrowd}*   & Arxiv21 & TRM & 66.1    & 95.4	 & 86	  \\
    	CCTrans \cite{CCTrans}*   & Arxiv21 & TRM & 64.4    & \textbf{95.4}	 & 104	  \\
		\hline
		\textbf{JCTNet(ours)*}	& ---	& CNN+TRM  & \textbf{62.8} & 95.6  & \textbf{28}	 \\
		\hline
	\end{tabular}
\end{table*}

\section{Related Work}
In this section, we briefly review some mainstream related works on crowd counting methods via fully supervised and weakly supervised. 
\subsection{Crowd Counting methods via fully supervised}
The crowd counting review \cite{CCsurvey} has introduced many excellent works.
Zhang et al. proposed a Multi-Column Convolution Neural Network, composed of three branches with convolution kernels of different sizes, namely MCNN \cite{MCNN}.
Later, Sam et al. proposed a density classifier, namely Switch CNN \cite{SwitchCNN}, for adaptively selecting optimal regressors from three different branches.
Boominathan et al. proposed CrowdNet \cite{CrowdNet}, including the shallow branch for extracting small-scale features and the deep branch for the opposite.
Guo et al. employed a multi-column dilated attention network to capture different receptive fields and focus on the crowd area \cite{DADNet}.
Later, Zhu et al. proposed a dual path fusion network, of which one path generated an attention map, and the other path generated the high-quality density map (SFANet \cite{SFANet}).
Sindagi et al. employed a hierarchical attention-based network, including a spatial attention module and several global attention modules (HACNN \cite{HACNN}).
Jiang et al. proposed attention scaling for Crowd Counting, namely ASNet \cite{ASNet}, composed of density attention network (DANet) for generating attention masks of different density levels and attention scaling network (ASNet) for generating several intermediate density maps and obtaining the final density maps.

\subsection{Crowd Counting methods via weakly supervised}
Lei et al. proposed the model MATT \cite{MATT} for weakly supervised crowd counting with a small amount of location-level annotations and a large amount of count-level annotations.
Yang et al. \cite{yang2020weakly} proposed the sorting network for directly regressing counting without location-level annotations. 
Sindagi et al. \cite{HACNN} proposed the weakly supervised model using image-level labels for adapting existing crowd counting models to new scenes and datasets.
The method \cite{unsupervised} proposed an almost unsupervised counting method for crowd counting.
However, the above works based on CNN did not achieve comparable counting performance.

Recently, the transformer has achieved significant performance by modeling the global interactions between different regions in many vision tasks, such as image classification \cite{VIT,Localvit}, image segmentation \cite{VIT,SwinTransformer,Twins,SwinUnet}, object detection \cite{EndtoendObject}.
Some recent works introduce the vision transformer into weakly supervised crowd counting.
Liang et al. first proposed a transformer-based weakly supervised crowd counting network, namely TransCrowd \cite{Transcrowd}, direct regression of the number of people by the proposed TransCrowd-Token and TransCrowd-GAP.
Tian et al. proposed simplifying and improving crowd counting with Transformer, namely CCTrans \cite{CCTrans}, which covered both full supervision and weak supervision method.

However, the above works usually employ the transformer as the backbone of the model.
This involves too many parameters and calculations and is not conducive to future lightweight research.

\section{Proposed Approaches}
In this section, we first illustrate the overview architecture of our proposed JCTNet for weakly supervised counting and then introduce each component in detail. 
The JCTNet includes three modules: (1) CNN feature extraction module (CFM); (2) Transformer feature extraction module (TRM); (3) counting regression module, as shown in Fig. 3.

\begin{figure*}[!t]
	\centering
	\includegraphics[width=7.0in]{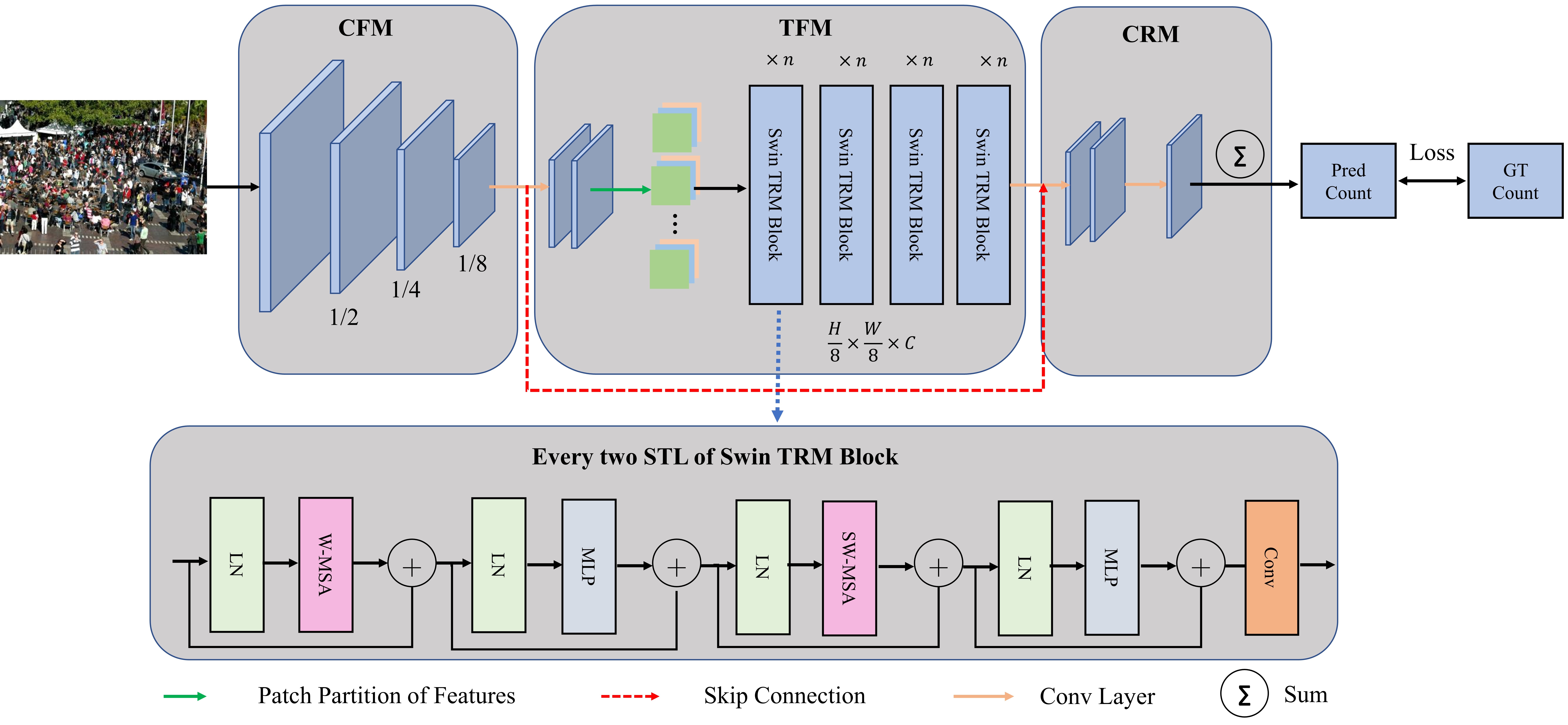}
	\caption{The proposed JCTNet overview. First, the input image is fed into the CNN feature extraction module (CFM) to learn crowd semantic features. 
	Then, by patch partition, these features are split into the 1D sequences, which go through the Transformer feature extraction module (TFM) for modeling global context. 
	Finally, the outputs of TFM are reshaped into 2D feature maps, and the counting regression module (CRM) serves for predicting the counting result. 
	STL denotes the Swin Transformer layer.}
	\label{fig_3}
\end{figure*}

\subsection{CNN feature extraction module}
The motivation for designing CFM is as follows: (1) Some transformer works \cite{VIT,SwinTransformer,Twins} directly split the input images into many patches then feed them into Transformer blocks.
The process is not appropriate for the individuals in crowd counting.
(2) For weakly supervised learning, in addition, deeper crowd semantic features extracted by CNN can bring better performance than shallow features such as color, texture, etc.
Hence, as shown in Fig. 3, we adopt the first ten layers of the VGG16-BN \cite{VGG} for crowd semantic feature extraction and as the input of the sequent TFM.

The CFM employs VGG16 as the extractor due to its excellent performance on many computers vision tasks such as object detection \cite{FasterRcnn}, classification \cite{Repvgg}, crowd counting \cite{Wnet}, etc.
Besides, W-net \cite{Wnet} has also demonstrated that VGG-BN could generate superior performance as the backbone of crowd counting.
To enlarge the receptive field while maintaining the number of parameters, we eliminate the last two pooling layers and ensure that the output of the network is 1/8 of the original image resolution.
Given a crowd input image $I \in {\mathbb{R}^{H \times W \times 3}}$ ($H$, $W$, and 3 are respectively its height, width, and channel size), the definition of CFM is as follows:

\begin{equation}
	{C_f} = {{\cal F}_{vgg}}(I)
\end{equation}

where $C_f$ is the feature from CNN, and ${\cal F}_{vgg}$ denotes the VGG16-BN first ten layers.

\subsection{Transformer feature extraction module}
In this module, we introduce the proposed Transfomer feature extraction module (TFM) using the output of CFM as the input.
Now, describe the data flow of our model as illustrated in Fig. 3.

\textbf{2D Feature to 1D Sequence.} Obtain the crowd semantic features $C_f \in {\mathbb{R}^{H \times W \times C}}$ from the above CFM, and then split it into $\frac{{HW}}{{{K^2}}}$ image patches and the size of each patch is $K \times K \times C$. 
Flatten these 2D patches features into the 1D sequence token $x \in {\mathbb{R}^{N \times D}}$, where $N = \frac{{HW}}{{{K^2}}}$, $D = K \times K \times C$.
We project a learnable projection $f:{x_i} \to {e_i} \in {\mathbb{R}^D}$ for mapping the $x$ into embedded features, i.e., image tokens.

\textbf{Modified Swin-Transformer Blocks.} We adopt modified Swin-Transformer \cite{SwinTransformer} as the primary transformer feature extractor due to its success in local attention and cross-window attention.
The TFM contains four modified Swin Transformer blocks.
In particular, we discard the patch merging layer to avoid the down-sample because the CFM has reduced the image’s resolution to 1/8 of the original input.
We add a convolution layer at the end of every two Swin Transformer layers of each modified Swin Transformer block to enhance the interaction between CNN and transformer features and bring the inductive bias of convolution into the transformer, motivated by SwinIR \cite{SWinIR}.
And, we add the skip connection for feature fusion between the CFM and the TFM.
The output of TFM is defined by the following equation:

\begin{equation}
	{T_F} = {{\cal F}_{TFM}}(Conv({C_f})) + Conv({C_f})
\end{equation}

where ${\cal F}_{TFM}$ are the Transformer feature extraction module, and ${T_F}$ denotes the 
obtained transformer features from TFM. $Conv$ is the convolution layer at the end of the CFM for channel dimension reduction and skip connection.

The TFM consists of 4 MSTB and each MSTB includes several Swin Transformer layers (STL). The workflow of MSTB is formulated as follows:

\begin{equation}
	{T_{{F_i}}} = {{\cal F}_{MST{B_i}}}({T_{{F_{i - 1}}}}),{\rm{ }}i = 1,{\rm{ }}2,{\rm{ }}3,{\rm{ }}4
\end{equation}
\begin{equation}
\begin{aligned}
\begin{split}
	{T_{{F_{i,j}}}} &= Conv({{\cal F}_{ST{L_{i,j}}}}({{\cal F}_{ST{L_{i,j - 1}}}}({T_{{F_{i,j - 2}}}}))),\\j &= 2,{\rm{ }}4,{\rm{ 6, 8, }} \ldots {\rm{, }}L
\end{split}
\end{aligned}
\end{equation}

where ${\cal F}_{MST{B_i}}$ are the $i$-th MSTB, and $T_{{F_i}}$ denotes the output features of each MSTB.
In Eq. (4), ${\cal F}_{ST{L_{i,j}}}$ is the $j$-th Swin Transformer layer of the $i$-th MSTB, and $T_{{F_{i,j}}}$ is the intermediate features by $L$ Swin Transformer layers. Specially, we combine two Swin Transformer layers and one convolution layer into a whole.

\textbf{Swin-Transformer layer.} Based on the standard transformer module \cite{VIT}, Swin Transformer \cite{SwinTransformer} proposed the window multi-head self-attention (W-MSA) for reducing the amount of calculation.
However, the window-based self-attention lacked interaction across windows and limited its power in modeling global context.
Then, Swin Transformer introduced shifted-window multi-head self-attention \cite{SwinTransformer} (SW-MSA) for maintaining the efficient computation of non-overlapping windows.
As shown at the bottom of Fig. 3, a standard 2-layer Swin Transformer block is the consecutive combination of W-MSA and SW-MSA, and the other layers are kept the same.
Given the input sequence ${Z_{l - 1}} \in {\mathbb{R}^{N \times D}}$ of $l$-th layer, first, reshape it into the 2D feature $H \times W \times C$. 
These features can be fed into W-MSA for computing local attention and sequent be transmitted into SW-MSA to execute cross-window connections.
The whole process is formulated as follows:

\begin{equation}
 	{Z_{l}^{'}}=WMSA(LN({Z_l}))+{Z_l}
\end{equation}
\begin{equation}
	{Z_{l}}=MLP(LN({Z_{l}^{'}}))+{Z_{l}^{'}}
\end{equation}
\begin{equation}
	{Z_{l+1}^{'}}=SWMSA(LN({Z_l}))+{Z_l}
\end{equation}
\begin{equation}
	{Z_{l+1}}=MLP(LN({Z_{l+1}^{'}}))+{Z_{l+1}^{'}}
\end{equation}
\begin{equation}
	{Z_{l+1}}=Conv(Z_{l+1})
\end{equation}

where $Z_{l}$ and $Z_{l+1}$ are the output features of the consecutive W-MSA and SW-MSA transformer blocks, respectively.
The MLP consists of two fully-connected layers with GELU non-linearity.
And, the LN denotes the layer normalization, which is added before both MSA and MLP.
The residual connection is employed for both modules.
Finally, add a convolution layer for features interaction between CNN and transformer features.

\subsection{Counting regression module}
The output of the above TFM is fed to the counting regression module (CFM) to predict the final count through a set of convolution layers.
These architectures are designed as follows: {Conv (256,128,3,2,2)-BN-ReLU, Conv (128,64,3,2,2)-BN-ReLU, Conv (64,1,1)}, where Conv (*) contains input channel, output channel, kernel size, dilation rate, padding, respectively.
The dilation rate is employed for enlarging receptive fields while containing the spatial resolution.

\section{Implementation Details}
In this section, we first introduce the five mainstream datasets. 
Sequentially, the loss function, training details are given respectively.

\subsection{Datasets}
\textbf{ShanghaiTech \cite{MCNN}} dataset contains 1,198 images with 330,165 annotated heads, divided into two parts: Part A and Part B.
Part A includes 300 training images and 182 test images randomly downloaded from the Internet, where the resolutions of each image are considerably different.
Part B is composed of 400 training images and 316 test images taken from streets in Shanghai, and the resolution of each image is 768$\times$1024.

\textbf{NWPU-Crowd \cite{NWPU}}, a large-scale and challenging dataset, consists of 5,109 images, 2,133,375 annotated heads with points and boxes in total, where these images are split into a training set (3,109), validation set (500), testing (1,500), respectively.
The dataset has obvious strengths: negative samples, online testing for fair evaluation, higher resolution, most annotated head, and large appearance variation, compared with previous datasets in the real world.

\textbf{UCF-CC-50 \cite{UCF_CC}} only contains 50 images of intensely congested scenes, a very challenging crowd counting dataset, during a total of 63,974 head annotations.
The number of pedestrians in each image ranges from 94 to 4,543, with an average number of 1,280 persons.
We perform 5-fold cross-validation by randomly selecting images to train and test our proposed approach due to its small size.

\textbf{UCF-QNRF \cite{QNRF}} contains 1,535 images from the Internet with 1,251,642 annotations, where it is divided into train and test sets of 1,201 and 334 images, respectively.
The number of pedestrians in each image varies from 49 to 12,865, with an average count of 815.4.
Specifically, the image resolutions and scale varied dramatically comparing other datasets.

\subsection{Loss Function}
We utilize SmoothL1 loss to measure the difference between predicted count and ground truth count.
Compared with L1, Smooth L1 is differentiable at the zero point and more robust to outliers.
It is defined as:

\begin{equation}
\begin{small}
	{{\cal L}} = \frac{1}{N}\sum\limits_{i = 1}^N {\left\{ {\begin{array}{*{20}{c}}
				{0.5{{(C_i^{ES} - C_i^{GT})}^2},{\rm{ }}|C_i^{ES} - C_i^{GT}| < 1}\\
				{|C_i^{ES} - C_i^{GT}| - 0.5,{\rm{ }}otherwise}
		\end{array}} \right.}
\end{small}
\end{equation}

where $N$ denotes the total number of images, and $C_i^{ES}$, $C_i^{GT}$ represent the estimated count and ground truth count, respectively.

\subsection{Training details}
All experimental training and evaluation are implemented on the platform of PyTorch \cite{Pytorch} with a GeForce RTX 3090 GPU.
The CNN feature extraction module (CFM) of the proposed JCTNet is leveraged from the first ten layers of VGG16-BN \cite{VGG}.
The modified Swin Transformer blocks inspired by the \cite{SwinTransformer} are used as the Transformer feature extraction module (TFM).
The hyper-parameters of the TFM are set as follows:

\begin{itemize}
	\item[$\bullet$] TFM: embed\_dim = 256, window\_size = 4, depths = [8, 8, 8, 8], num\_heads = [8, 8, 8, 8],
\end{itemize}

where embed\_dim is the embedding dimension of hidden layers, depths is the depth of each modified Swin Transformer block, and num\_heads is the number of attention heads in different layers.
In the training phase, we randomly crop image patches with the size of m$\times$n pixels from the original image to ensure our network can be multi-batch trained and promote performance at a lower time cost.
The m,n configuration is detailed in Table 2.
We use AdamW \cite{AdamW} optimizer with a learning rate of 1e-5 and weight decay of 1e-4 to train our model by minimizing the loss function Eq. (10).
The batch size is set to 8 or 16 according to GPU computing power and the number of iterations is set to 2000.

\begin{table}[!t]
	\centering
	\caption{The crop size of image patches.}
	\label{Table 2}
	\begin{tabular}{|c|c|}
		\hline
		Datasets & (m, n)  \\ \hline
		ShanghaiPart A  & (256, 256) \\ \hline
		ShanghaiPart B  & (512, 512) \\ \hline
		UCF-CC-50  & (256, 256) \\ \hline
		UCF-QNRF  & (512, 512) \\ \hline
		NWPU-Crowd  & (384, 384) \\ \hline
	\end{tabular}
\end{table}

\section{Experiments and Results}
In this section, we display the evaluation metrics and compare the results of the proposed JCTNet with state-of-the-art methods. 
In the end, we perform extensive ablation experiments to validate the effectiveness of our method.

\subsection{Evaluation metrics}
To evaluate the accuracy of our approach, the mean absolute error (MAE), the mean squared error (MSE), and the mean normalized absolute error (NAE) \cite{NWPU} are adopted as metrics.
Specifically, equations are defined as:

\begin{equation}
	MAE = \frac{1}{N}\sum\nolimits_{i = 1}^N {|C_i^{ES} - C_i^{GT}|}
\end{equation}

\begin{equation}
	MSE = {(\frac{1}{N}\sum\nolimits_{i = 1}^N {|C_i^{ES} - C_i^{GT}{|^2}} )^{\frac{1}{2}}}
\end{equation}

\begin{equation}
	NAE = \frac{1}{N}\sum\nolimits_{i = 1}^N {\frac{{|C_i^{ES} - C_i^{GT}|}}{{C_i^{GT}}}}
\end{equation}

where $N$ is the number of test images. $C_i^{ES}$ is the estimated count of the $i$-th image, and $C_i^{GT}$ is the corresponding ground truth count of the $i$-th image.

\subsection{Comparisons with state-of-the-art}
The elaborate comparisons and visual presentation are conducted with state-of-the-art on five benchmark datasets to demonstrate the effectiveness of the proposed weakly-supervised crowd counting approach JCTNet.
We compare the proposed method with the existing fully-supervised methods (via location-level annotations) and weakly-supervised methods (via count-level annotations).

\begin{table*}[ht]
	\centering
	\caption{Estimation errors on the ShanghaiTech dataset. * represents the weakly-supervised method.}
	\setlength{\tabcolsep}{3pt}
	\label{Table 3}
	\begin{tabular}{|c|c|cc|cc|cc|}
		\hline
	    \multirow{2}{*}{Method} & \multirow{2}{*}{Publication}  & \multicolumn{2}{c|}{Label} & \multicolumn{2}{c|}{Part A}   & \multicolumn{2}{c|}{Part B}  \\ \cline{3-8} 
		{}  & {}     & Location  &Count         & MAE           & MSE           & MAE          & MSE           \\ \hline
		SANet \cite{SANet}   & ECCV18  & \Checkmark  & \Checkmark   & 67.0    & 104.5    & 8.4      & 13.6          \\
		SFCN+ \cite{SFCN}    & CVPR19  & \Checkmark  & \Checkmark   & 64.8    & 107.5    & 7.6      & 13.0          \\
		TEDNet \cite{TEDNet}    & CVPR19  & \Checkmark  & \Checkmark   & 64.2    & 109.1    & 8.2    & 12.8        \\
		DADNet \cite{DADNet}    & MM19  & \Checkmark  & \Checkmark   & 64.2    & 99.8    & 8.8        & 13.5        \\
		Density CNN \cite{DensityCNN}    & TMM19  & \Checkmark  & \Checkmark   & 63.1    & 106.3    & 9.1    & 16.3    \\
		HACNN \cite{HACNN}    & TIP19  & \Checkmark  & \Checkmark   & 62.9    & 94.9    & 8.1      & 13.4        \\
		CAN \cite{CAN}    & CVPR19  & \Checkmark  & \Checkmark   & 62.3    & 100.0    & 7.8      & 12.2        \\ 
		PaDNet \cite{PaDNet}    & TIP19  & \Checkmark  & \Checkmark   & 59.2    & 98.1    & 8.1      & 12.2        \\
		PGCNet \cite{PGCNet}    & ECCV19  & \Checkmark  & \Checkmark   & 57.0    & 86.0    & 8.8      & 13.7        \\
		DM-Count \cite{DM-Count}    & NeurIPS20  & \Checkmark  & \Checkmark   & 59.7    & 95.7    & 7.4      & 11.8    \\
		ASNet \cite{ASNet}    & CVPR20  & \Checkmark  & \Checkmark   & 57.7    & 90.1    & ---      & ---    \\
		P2PNet \cite{P2PNet}    & ICCV21  & \Checkmark  & \Checkmark   & 52.7    & 85.1    & 6.3      & \textbf{9.9}    \\
		BCCT \cite{BCCT}    & Arxiv21  & \Checkmark  & \Checkmark   & 53.1    & \textbf{82.2}    & 7.3      & 11.3    \\
		CCTrans \cite{CCTrans}    & Arxiv21  & \Checkmark  & \Checkmark   & \textbf{52.3}    & 84.9    & 6.2   & \textbf{9.9}  \\
	    \hline
	    Yang et al. \cite{yang2020weakly}*   & ECCV20  & \XSolid  & \Checkmark   & 104.6    & 145.2    & 12.3     & 21.2    \\
	    MATT \cite{MATT}*   & PR21  & \XSolid  & \Checkmark   & 80.1    & 129.4    & 11.7     & 17.5    \\
	    TransCrowd \cite{Transcrowd}*   & Arxiv21  & \XSolid  & \Checkmark   & 66.1    & 105.1    & 9.3   & 16.1    \\
	    CCTrans \cite{CCTrans}*   & Arxiv21  & \XSolid  & \Checkmark   & 64.4  & \textbf{95.4}  & \textbf{7.0}  & \textbf{11.5}    \\
	    JCTNet(ours) *   & ---   & \XSolid  & \Checkmark   & \textbf{62.8}    & 95.6    & 7.2  & \textbf{11.5}    \\
	    \hline
	\end{tabular}
\end{table*}

\textbf{Result on ShanghaiTech.} To evaluate the effectiveness of the proposed JCTNet, we compare the model with existing state-of-the-art 14 fully-supervised and 4 weakly-supervised methods on the ShanghaiTech dataset.
As shown in Table 3, compared with the weakly-supervised crowd counting methods, the proposed JCTNet can achieve the lowest MAE of 62.8 and a comparable MSE of 95.6 on Part A and obtain the second places MAE of 7.2 and MSE of 11.5 on Part B.
Compared with the fully-supervised crowd counting methods, our model still obtains relatively outstanding performance, though it is unfair to measure the fully-supervised methods and weakly-supervised methods.

\begin{table}[ht]
	\centering
	\caption{Estimation errors on the UCF-QNRF dataset. * represents the weakly-supervised method.}
	\setlength{\tabcolsep}{3pt}
	\label{Table 4}
	\begin{tabular}{|c|c|cc|cc|}
		\hline
		\multirow{2}{*}{Method} & \multirow{2}{*}{Publication}  & \multicolumn{2}{c|}{Label} & \multicolumn{2}{c|}{UCF-QNRF}  \\ \cline{3-6} 
		{}  & {}     & Location  &Count         & MAE           & MSE   \\ 
		\hline
		SFCN+ \cite{SFCN}    & CVPR19  & \Checkmark  & \Checkmark   & 102    & 171       \\
		TEDNet \cite{TEDNet}    & CVPR19  & \Checkmark  & \Checkmark   & 113    & 188   \\
		DADNet \cite{DADNet}    & MM19  & \Checkmark  & \Checkmark   & 113    & 189     \\
		Density CNN \cite{DensityCNN}    & TMM19  & \Checkmark  & \Checkmark   & 101    & 186    \\
		CAN \cite{CAN}    & CVPR19  & \Checkmark  & \Checkmark   & 107    & 183    \\ 
		PaDNet \cite{PaDNet}    & TIP19  & \Checkmark  & \Checkmark   & 96    & 170    \\
		DM-Count \cite{DM-Count}    & NeurIPS20  & \Checkmark  & \Checkmark   & 85    & 148    \\
		ASNet \cite{ASNet}    & CVPR20  & \Checkmark  & \Checkmark   & 91    & 159   \\
		P2PNet \cite{P2PNet}    & ICCV21  & \Checkmark  & \Checkmark   & 85    & 154    \\
		BCCT \cite{BCCT}   & Arxiv21  & \Checkmark  & \Checkmark   & 83    & 143    \\
		CCTrans \cite{CCTrans}    & Arxiv21  & \Checkmark  & \Checkmark   & \textbf{82}    & \textbf{142}    \\
		\hline
		TransCrowd \cite{Transcrowd}*   & Arxiv21  & \XSolid  & \Checkmark   & 97    & 168    \\
		CCTrans \cite{CCTrans}*   & Arxiv21  & \XSolid  & \Checkmark   & 92  & \textbf{158}  \\
		JCTNet(ours) *   & ---   & \XSolid  & \Checkmark   & \textbf{90}    & 161   \\
		\hline
	\end{tabular}
\end{table}

\textbf{Result on UCF-QNRF.} Compared with other weakly supervised works, our model obtains the best MAE result with 90 and the second best MSE with 161.
Besides, The model parameters of TransCrowd and CCTrans \cite{CCTrans} are 86M and 104M, respectively, in Table 1, while the proposed JCTNet parameters are only 28M, achieving a reduction of 67\% and 73\%.
Compared with the fully supervised works, our model still obtains significant counting results.

\begin{table}[ht]
	\centering
	\caption{Estimation errors on the UCF-CC-50 dataset. * represents the weakly-supervised methods.}
	\setlength{\tabcolsep}{3pt}
	\label{Table 5}
	\begin{tabular}{|c|c|cc|cc|}
		\hline
		\multirow{2}{*}{Method} & \multirow{2}{*}{Publication}  & \multicolumn{2}{c|}{Label} & \multicolumn{2}{c|}{UCF-CC}  \\ \cline{3-6} 
		{}  & {}     & Location  &Count         & MAE           & MSE   \\ 
		\hline
		SANet \cite{SANet}    & ECCV18  & \Checkmark  & \Checkmark   & 377.6    & 509.1     \\
		SFCN+ \cite{SFCN}     & CVPR19  & \Checkmark  & \Checkmark   & 214.2    & 318.2       \\
		TEDNet \cite{TEDNet}    & CVPR19  & \Checkmark  & \Checkmark   & 249.4    & 354.5   \\
		Density CNN \cite{DensityCNN}    & TMM19  & \Checkmark  & \Checkmark   & 244.6   & 341.7    \\
		CAN \cite{CAN}    & CVPR19  & \Checkmark  & \Checkmark   & 212.2   & 243.7   \\ 
		DM-Count \cite{DM-Count}    & NeurIPS20  & \Checkmark  & \Checkmark   & 211.0    & 291.5   \\
		ASNet \cite{ASNet}    & CVPR20  & \Checkmark  & \Checkmark   & 174.8    & 251.6   \\
		P2PNet \cite{P2PNet}    & ICCV21  & \Checkmark  & \Checkmark   & 172.7   & 256.2    \\
		CCTrans \cite{CCTrans}    & Arxiv21  & \Checkmark  & \Checkmark   & \textbf{168.7}    & \textbf{234.5}    \\
		\hline
		MATT \cite{MATT}*   & PR21  & \XSolid  & \Checkmark   & 355.0    & 550.2    \\
		CCTrans \cite{CCTrans}*   & Arxiv21  & \XSolid  & \Checkmark   & 245.0  & 343.6  \\
		JCTNet(ours) *   & ---   & \XSolid  & \Checkmark   & \textbf{222.9}    & \textbf{306.5}   \\
		\hline
	\end{tabular}
\end{table}

\textbf{Result on UCF-CC-50.} This is an extremely ingested crowd but small dataset, so we perform a 5-fold cross-validation for fairly evaluating the capability of the JCTNet.
As shown in Table 5, the proposed method achieves the best result with an MSE of 222.9 and MSE of 306.5, compared with other weakly supervised methods.

\begin{table*}[ht]
	\centering
	\caption{Estimation errors on the NWPU-Crowd dataset. * represents the weakly-supervised methods.}
	\setlength{\tabcolsep}{3pt}
	\label{Table 6}
	\begin{tabular}{|c|c|cc|cc|ccc|}
		\hline
		\multirow{2}{*}{Method} & \multirow{2}{*}{Publication}  & \multicolumn{2}{c|}{Label} & \multicolumn{2}{c|}{Val}   & \multicolumn{3}{c|}{Test}  \\ \cline{3-9} 
		{}  & {}     & Location  &Count         & MAE           & MSE           & MAE          & MSE    & NAE        \\ 
		\hline
		MCNN \cite{MCNN}    & CVPR16  & \Checkmark  & \Checkmark   & 218.5    & 700.6    & 232.5    & 714.6  & 1.063      \\
		SANet \cite{SANet}   & ECCV18  & \Checkmark  & \Checkmark   & 171.1    & 471.5    & 190.6   & 491.4  & 0.991      \\
		CSRNet \cite{CSRNet}    & CVPR18  & \Checkmark  & \Checkmark   & 104.9    & 433.4    & 121.3  & 387.8  & 0.604      \\
		SFCN+ \cite{SFCN}    & CVPR19  & \Checkmark  & \Checkmark   & 95.5    & 608.3    & 105.7    & 424.1   & 0.254      \\
		CAN \cite{CAN}    & CVPR19  & \Checkmark  & \Checkmark   & 93.6    & 489.9    & 106.3      & 386.5   & 0.295     \\ 
		BL \cite{BLoss}    & ICCV19  & \Checkmark  & \Checkmark   & 93.6    & 470.4    & 105.4     & 454.2 & 0.203      \\
		DM-Count \cite{DM-Count}    & NeurIPS20  & \Checkmark  & \Checkmark   & 70.5    & 357.6    & 88.4   & 388.6  & 0.169   \\
		KDMG \cite{KDMG}   & TPAMI20  & \Checkmark  & \Checkmark   & ---    & ---    & 100.5    & 415.5  & ---    \\
		NoisyCC \cite{wan2020modeling} & NeurIPS20  & \Checkmark  & \Checkmark  & ---    & ---   & 96.9    & 534.2   & ---    \\
		P2PNet \cite{P2PNet}    & ICCV21  & \Checkmark  & \Checkmark   & ---    & ---    & 77.4      & 362.0 & ---   \\
		BCCT \cite{BCCT}    & Arxiv21  & \Checkmark  & \Checkmark   & 53.0    & 170.3    & 82.0    & 366.9  & 0.164   \\
		CCTrans \cite{CCTrans}    & Arxiv21  & \Checkmark  & \Checkmark & \textbf{38.6}  & \textbf{87.8} & \textbf{69.3}   & \textbf{299.4} & \textbf{0.135} \\
		\hline
		TransCrowd \cite{Transcrowd}*   & Arxiv21  & \XSolid  & \Checkmark   & 88.4    & 400.5    & 117.7   & 451.0 &  0.244  \\
		CCTrans \cite{CCTrans}*   & Arxiv21  & \XSolid  & \Checkmark   & \textbf{48.6}  & \textbf{121.1}  & \textbf{79.8}  & \textbf{344.4}  &  \textbf{0.157} \\
		JCTNet(ours) *   & ---   & \XSolid  & \Checkmark   & 68.1   & 310.6  & 83.4  & 385.0 & 0.185   \\
		\hline
	\end{tabular}
\end{table*}

\textbf{Result on NWPU-Crowd.} Table 6 presents the counting errors on the NWPU-Crowd dataset. 
The proposed JCTNet wins second place with an MAE of 83.4 and MSE of 385.0 compared with the weakly supervised methods.
The results of JCTNet are relatively poor compared with CCTrans \cite{CCTrans}, but it has about a quarter of the parameters of method CCTrans, only 28M as shown in Table 1.
We find that the proposed model performs poorly on large-scale images containing more pedestrian annotations, suggesting that models constrained by count-level annotations cannot better adapt to scale changes.
This will be the next research.

\subsection{Visualization presentation}
\textbf{Visualization of convergence curves.} Fig. 4 displays the convergence curves of the proposed JCTNet on the ShanghaiTech Part A dataset, including Smooth L1 loss, MAE, and MSE metrics, where the starting epoch of the three curves are from 0 to 2,000, 100 to 2,000, and 100 to 2,000, respectively.
There are two observations: (1) The training loss can drop rapidly and converge fast.
(2) The model has achieved the best counting performance at the 353-th epoch, i.e., MAE of 62.8 and MSE of 95.6.
This figure further demonstrates the effectiveness of the proposed model.

\begin{figure}[!t]
	\centering
	\includegraphics[width=2.5in]{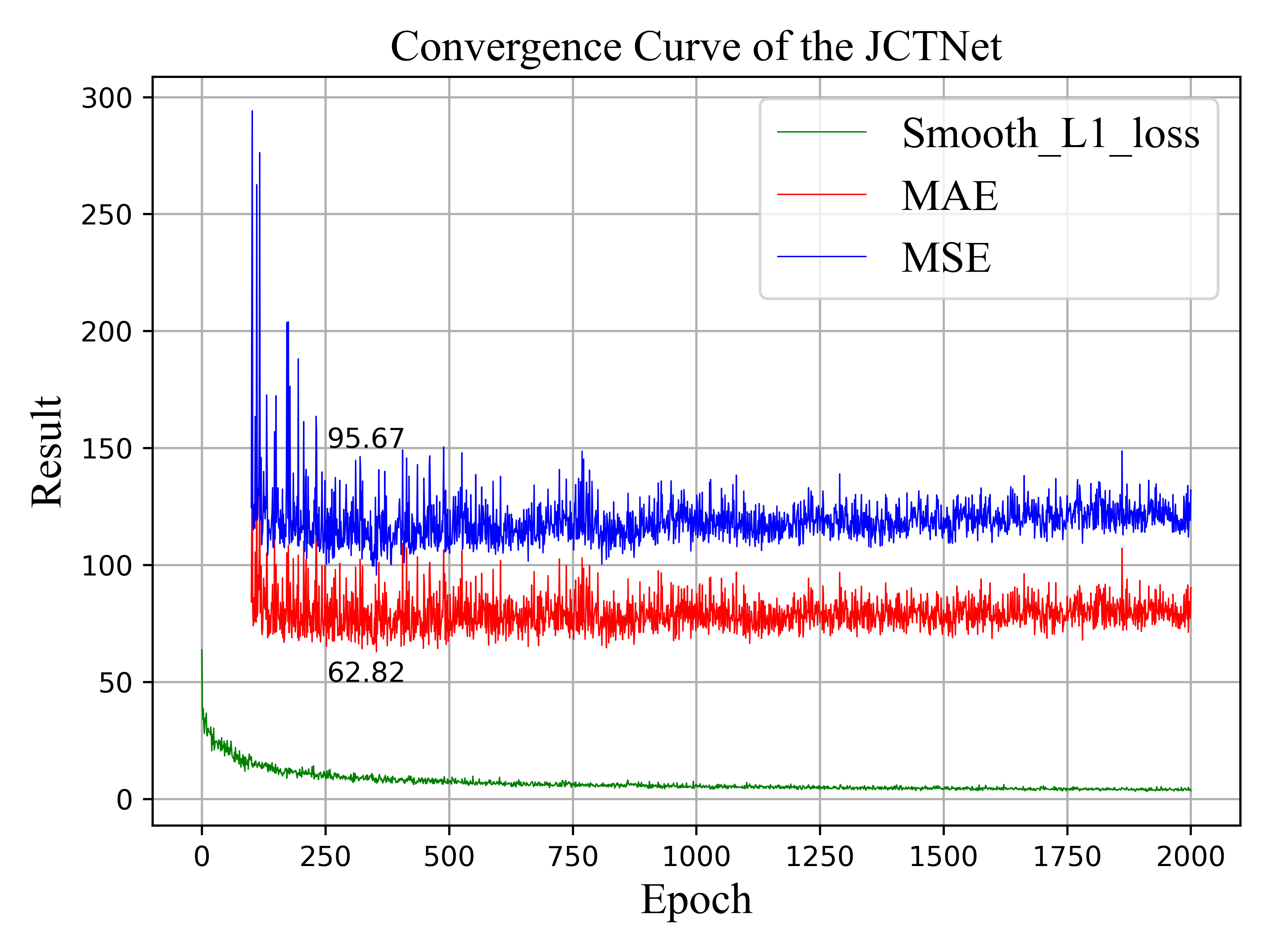}
	\caption{The convergence curves of Smooth L1 Loss, MAE, and MSE metrics of the proposed JCTNet on the ShanghaiTech Part A dataset.}
	\label{fig_4}
\end{figure}

\textbf{Visualization of the model’s components.} For powerfully exploring what the model attends to in weakly supervision crowd counting, we present qualitative visualizations of intermediate feature maps of the model, consisting of the output feature of CFM, TFM, and CRM.
Specially, for the features of size $H \times W \times C$ (height, width, channel), we select one from the $C$ feature maps for visualization, as shown in Fig. 5.
We observe that the TFM can successfully learn the contrastive features and distinguish foreground and background regions (in black rectangles), and to our surprise, CRM can significantly focus on crowd areas (in red rectangles) and get superior counting results only with count-level annotations.

\begin{figure*}[!t]
	\centering
	\includegraphics[width=5.0in]{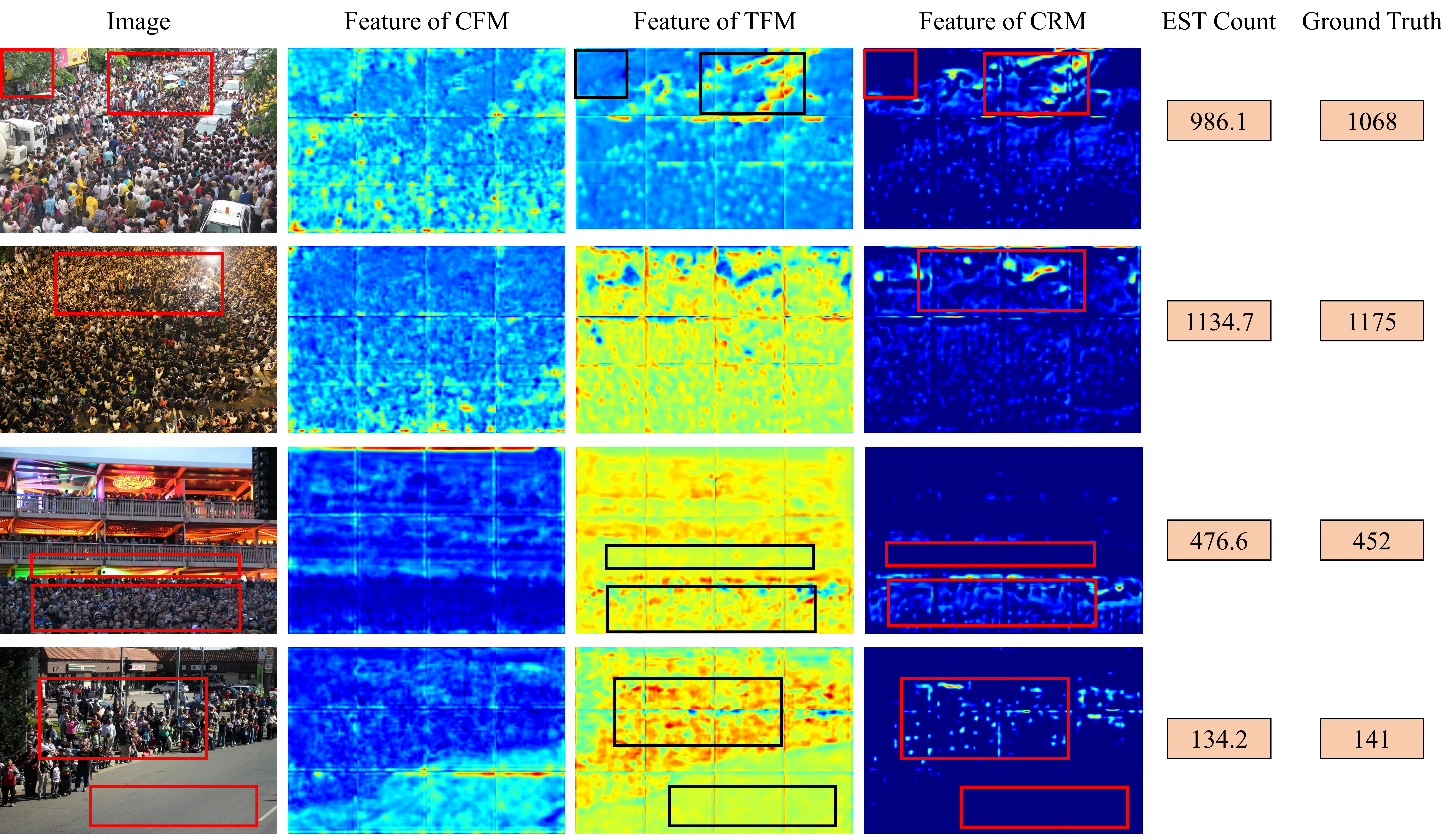}
	\caption{The visualization of the feature maps: the output feature of CFM, TFM, and CRM. For weakly supervised crowd counting, the TFM of the proposed JCTNet can be still able to distinguish crowd region and background (the black rectangle), and then the CRM focus on counting crowd region (the red rectangle).}
	\label{fig_5}
\end{figure*}

\textbf{Visualization of the heat map of the last convolution layer of the proposed JCTNet.} In Fig. 6, we give the high-quality heat map of the last convolution layer of the proposed JCTNet to further understand what the model prefers to focus on in the weakly supervision crowd counting, inspired by \cite{GradCam}.
We observe that although the model is only weakly supervised by count-level annotations, it is still able to pay close attention to crowd regions.

For the above phenomenon, we give a possibility analysis: (1) Transformer component can sufficiently learn the contrasting features in favor of distinguishing various objects.
(2) Then the CNN can count various objects with similar characteristics, respectively, due to its significant inductive bias capability.
(3) Finally, the model regards the category whose count result is closest to the ground truth as the feature to take into count, i.e., the target feature.
We hope this exploration will spur further research in the field.

\begin{figure}[!t]
	\centering
	\includegraphics[width=3.0in]{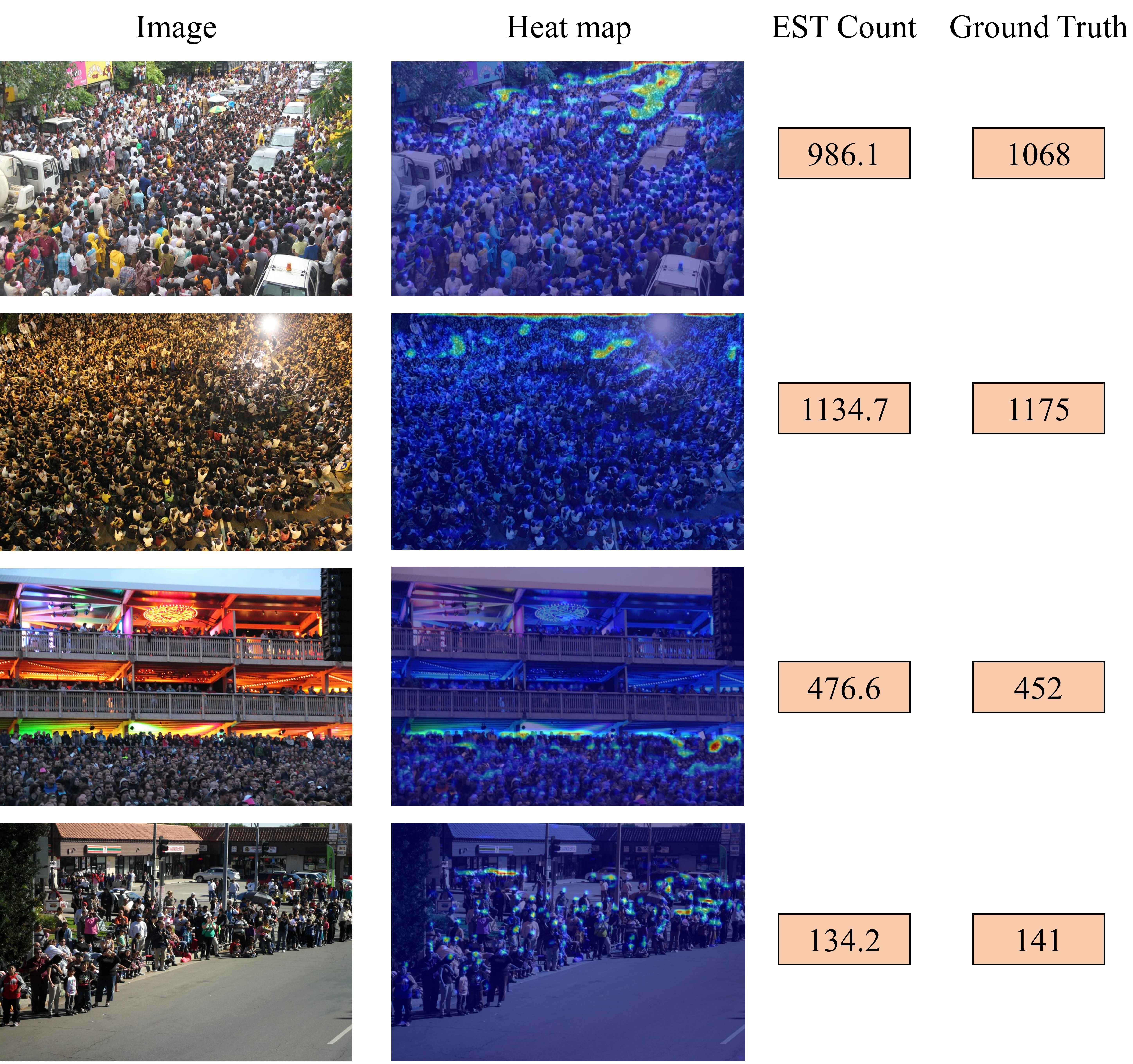}
	\caption{The visualization of the heat map of the last convolution layer of the proposed JCTNet on input images: It can be observed that the model is still able to focus the crowd regions for weakly supervised crowding counting.}
	\label{fig_6}
\end{figure}

\subsection{Ablation Experiments}
In this section, we conduct multigroup ablation experiments to evaluate the effectiveness of different components on the ShanghaiTech Part A dataset.

\textbf{Ablation study on model’s components.} Table 7 presents the ablation study of the component of the proposed JCTNet on the ShanghaiTech Part A dataset. 
From the table, CFM, TFM, and CRM provide progressive improvements, respectively.
Specifically, CFM achieves the MAE of 77.6 and the MSE of 127.5, and by adding TFM over CFM, we observe performance improvement of 13.4 and 21.8 on MAE and MSE, which demonstrates the effectiveness of the joint learning between CNN and Transformer.
By further adding CRM, the best results are obtained.

\begin{table}[ht]
	\centering
	\caption{Ablation study on key components of JCTNet.}
	\setlength{\tabcolsep}{3pt}
	\label{Table 7}
	\begin{tabular}{|c|cc|cc|}
		\hline
		\multirow{2}{*}{Method} & \multicolumn{2}{c|}{Label}   & \multicolumn{2}{c|}{Part A}  \\ \cline{2-5} 
		{}                 & Location           & Count           & MAE          & MSE           \\ 
		\hline
		CFM           & \XSolid  & \Checkmark        & 77.6         & 127.5          \\
		CFM + CRM     & \XSolid  & \Checkmark        & 75.0          & 126.3          \\
	    CFM + TFM     & \XSolid  & \Checkmark        & 64.2          & 105.7          \\
	    CFM + TFM + CRM    & \XSolid  & \Checkmark        & \textbf{62.8}          & \textbf{95.6}      \\
		\hline
	\end{tabular}
\end{table}

\textbf{Ablation study on TFM's embedding dimension and Swin Transformer Layer (STL) number of each Modified Swin Transformer Block (MSTB).}
Table 8 the ablation study of embedding dimension and STL number of each MSTB for the proposed JCTNet on ShanghaiTech A.
The x, y of the TFM (x, y) denote the embedding dimension and STL number of each MSTB, respectively.
From the table, by enlarging the embedding dimension and the number of STLs, the count error (MAE) decreases gradually as more parameters are involved.

\begin{table}[ht]
	\centering
	\caption{Ablation study on the embedding dimension and number of STL of TFM.}
	\setlength{\tabcolsep}{3pt}
	\label{Table 8}
	\begin{tabular}{|c|cc|cc|c|}
		\hline
		\multirow{2}{*}{Method} & \multicolumn{2}{c|}{Label}   & \multicolumn{2}{c|}{Part A}  & \multirow{2}{*}{\shortstack{Parameters\\ (M)}} \\ \cline{2-5} 
		{}   & Location    & Count         & MAE          & MSE   & {}    \\ 
		\hline
	    CFM + TFM (64, 8) + CRM           & \XSolid  & \Checkmark        & 66.3    & 111.2   & 10.4        \\
		CFM + TFM (128, 8) + CRM     & \XSolid  & \Checkmark      & 64.5     & 99.9   & 14.2     \\
		CFM + TFM (256, 8) + CRM     & \XSolid  & \Checkmark      & \textbf{62.8}     & \textbf{95.6}   & 28.8     \\
		CFM + TFM (256, 2) + CRM     & \XSolid  & \Checkmark      & 68.4     & 117.1   & 16.3     \\
		CFM + TFM (256, 4) + CRM     & \XSolid  & \Checkmark      & 65.7     & 102.8   & 20.5     \\
		CFM + TFM (256, 8) + CRM     & \XSolid  & \Checkmark      & \textbf{62.8}     & \textbf{95.6}   & 28.8     \\
		\hline
	\end{tabular}
\end{table}

\textbf{Ablation study on the Loss function.} We perform quantitative analysis between the L1 loss and Smooth L1 loss on the ShanghaiTech Part A, as shown in Table 9. 
The MAE and MSE can be reduced from 63.7 and 98.1 to 62.8 and 95.6 by employing the Smooth L1 loss.

\begin{table}[ht]
	\centering
	\caption{Ablation study on the loss function.}
	\setlength{\tabcolsep}{3pt}
	\label{Table 9}
	\begin{tabular}{|c|cc|cc|}
		\hline
		\multirow{2}{*}{Method} & \multicolumn{2}{c|}{Label}   & \multicolumn{2}{c|}{Part A}  \\ \cline{2-5} 
		{}                 & Location           & Count           & MAE          & MSE           \\ 
		\hline
		L1 Loss           & \XSolid  & \Checkmark        & 63.7         & 98.1          \\
		Smooth L1 Loss     & \XSolid  & \Checkmark        & \textbf{62.8}          & \textbf{95.6}         \\
		\hline
	\end{tabular}
\end{table}

\section{Conclusion}
In this paper, we proposed a joint CNN and Transformer learning network for the efficient weakly supervised crowd counting, named JCTNet. 
The model includes three parts: CNN feature extraction module (CFM) for extracting crowd semantic information, Transformer extraction module (TFM) for capturing the global context and thoroughly learning the contrast features between foreground and background, and counting regression module (CRM) for estimating the final count of people. 
Extensive experiments and visualization demonstrate the effectiveness of the proposed JCTNet on five mainstream datasets.

\section*{Acknowledgments}
This work was supported by National Natural Science Foundation of China (No. 61971073).


\bibliographystyle{IEEEtran}
\bibliography{bare_jrnl_new_sample4}

%
%
%
%
%

\vfill

\end{document}